\documentclass{article}

\usepackage{template/arxiv}

\usepackage{cite}
\usepackage{amsmath,amssymb,amsfonts}
\usepackage{algorithmic}
\usepackage{graphicx}
\usepackage{textcomp}
\usepackage{xcolor}
\def\BibTeX{{\rm B\kern-.05em{\sc i\kern-.025em b}\kern-.08em
    T\kern-.1667em\lower.7ex\hbox{E}\kern-.125emX}}

\usepackage[hidelinks]{hyperref}
    
\usepackage{accents}
\DeclareRobustCommand{\groupderiv}[1]{\accentset{\scriptstyle\circ}{#1}}

\newcommand{\BodyForce}{{F}}

\newcommand{\LinkNumber}{{i}}

\newcommand{\DragMatrix}{{C}}

\newcommand{\Jacobian}{{J}}

\newcommand{\Shape}{{\alpha}}
\newcommand{\ShapeChange}{{\dot\Shape}}
\newcommand{\Gait}{{\phi}}
\newcommand{\Time}{{t}}

\newcommand{\BodyFrame}{{g}}
\newcommand{\BodyCirc}{\groupderiv{g}}
\newcommand{\BodyPrime}{\BodyFrame'}

\newcommand{\LinkVelocity}{\BodyCirc_\LinkNumber}

\newcommand{\Motility}{{A}}
\newcommand{\MotilitySR}{\Motility_\text{SR}}
\newcommand{\MotilityRS}{\Motility_\text{RS}}
\newcommand{\MotilityPlus}{\Motility^+}
\newcommand{\MotilityMinus}{\Motility^-}
\newcommand{\MotilityDiff}{\Motility^{\pm}}

\newcommand{\LinkLength}{{L}}
\newcommand{\DragCoeff}{{\mu}}
\newcommand{\DragLat}{{\DragCoeff_\text{lat}}}
\newcommand{\DragLon}{\DragCoeff_\text{lon}}
\newcommand{\BWRatio}{{\DragLon^-}}
\newcommand{\FWRatio}{{\DragLon^+}}

\newcommand{\MotilityLeft}{$\triangleleft$ }
\newcommand{\MotilityRight}{$\textcolor{red}{\triangleright}$ }

\begin{document}

\title{Scales and Locomotion: Non-Reversible Longitudinal Drag
\thanks{This work was supported in part by the NSF under awards 1653220 and 1826446.}
}

\author{Quinten Konyn\\
Collaborative Robotics and\\
Intelligent Systems (CoRIS) Institute\\
Oregon State University\\
Corvallis, Oregon, USA \\
konynq@oregonstate.edu
\And
Ross L. Hatton\\
Collaborative Robotics and\\
Intelligent Systems (CoRIS) Institute\\
Oregon State University\\
Corvallis, Oregon, USA \\
ross.hatton@oregonstate.edu
}

\maketitle

\begin{abstract}
Locomotion requires that an animal or robot be able to move itself forward farther than it moves backward in each gait cycle (formally, that it be able to break the symmetry of its interactions with the world). Previous work has established that a difference between lateral and longitudinal drag provides sufficient conditions for locomotion to be possible. The geometric mechanics community has used this principle to build a geometric framework for describing the effectiveness and efficiency of undulatory locomotion. Researchers in biology and robotics have observed that structures such as snake scales additionally provide a difference between forward and backward longitudinal drag. As yet, however, the impact of scales on the geometric features relevant to locomotion effectiveness and efficiency have not yet been explored. We present a geometric model for a single-joint undulating system with scales and identify the features needed to understand its motion. Mathematically, the scales can be treated as inducing a "Finsler metric" on the configuration space, and this paper lays the groundwork for further research into application of such Finsler metrics to robotic locomotion.
\end{abstract}


\thispagestyle{plain}
\pagestyle{plain}

\section{Introduction}

To locomote, an animal or robot can deform its body in a pattern that interacts with the environment to create net motion.
For both biological research and robot design, it is important to develop an in-depth mathematical intuition for how these interactions work.

The dominant mechanism of body-environment interaction varies depending on the specific situation, but must take advantage of some form of asymmetry in order to pick out a direction of motion.
One well-understood source of asymmetry is body parts that are long and narrow, so that lateral drag is larger than longitudinal drag. Purcell's swimmer \cite{Purcell:1977} is a classical example. This kind of asymmetry\footnote{commonly referred to as anisotropy in the literature} is reversible: if the swimmer executes a gait and then does that same gait in reverse, it will end up back where it started.

\begin{figure}
    \centering
    \includegraphics[height=3in]{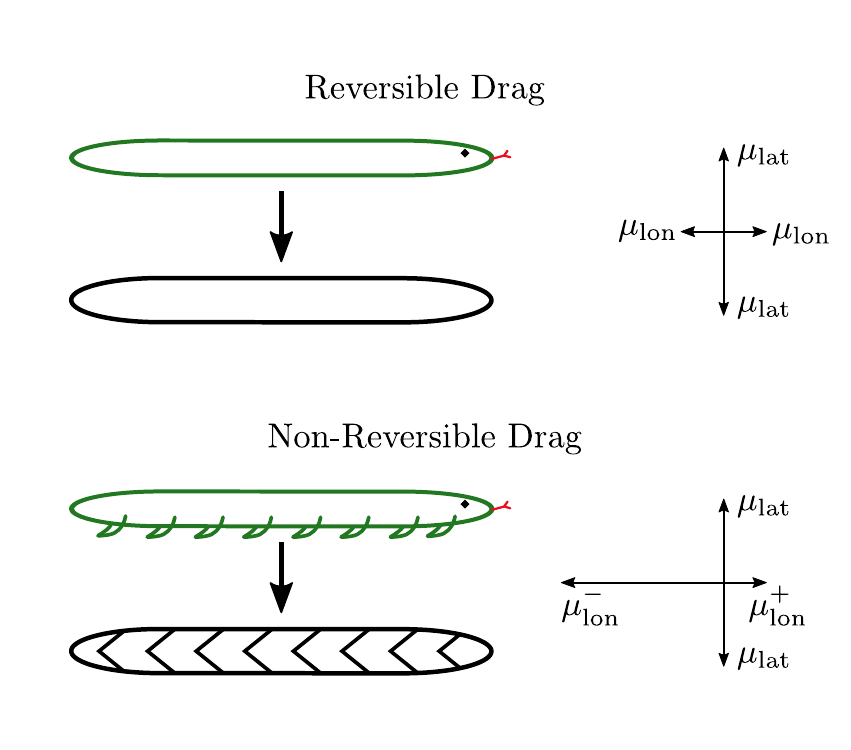}
    \caption{Without scales, we model longitudinal drag as smaller than lateral drag, $\DragLat = 2\DragLon$. Neither direction depends on the sign of the motion. However with scales, we model longitudinal drag as much larger when moving backward, $\BWRatio = 4\DragLon$ and unchanged when moving forward, $\FWRatio = \DragLon$.}
    \label{fig:KindsOfDrag}
\end{figure}

Many animals have scales that, in addition to making longitudinal drag smaller than lateral drag, make forward drag smaller than backward drag\cite{Hu:2009}\cite{Gray:1950}, as in Fig. \ref{fig:KindsOfDrag}. Experimental research shows that scales on the bottom of a snake robot make the robot move farther than without scales\cite{Branyan:2020}\cite{Serrano:2015}, but the theory behind non-reversible asymmetry\footnote{also commonly referred to as anisotropy in the literature, or sometimes longitudinal anisotropy} is not yet well understood.

This research develops a method to apply non-reversible systems with the existing tools of geometric mechanics\cite{Ramasamy:2019}.
We model the minimal locomoting non-reversible system, the scaled two-link swimmer, as a piecewise combination of two-link non-scaled swimmers. Each of these reversible swimmers is analogous to the scaled swimmer in specific situations, depending on the links' current body velocities, and each on its own is compatible with existing tools.
We then extract the key geometric features relevant to the net displacement over a gait cycle, emerging from the piecewise model with a deeper understanding of the scaled system as a whole.

\begin{figure*}
    \centering
    \includegraphics[width=\textwidth]{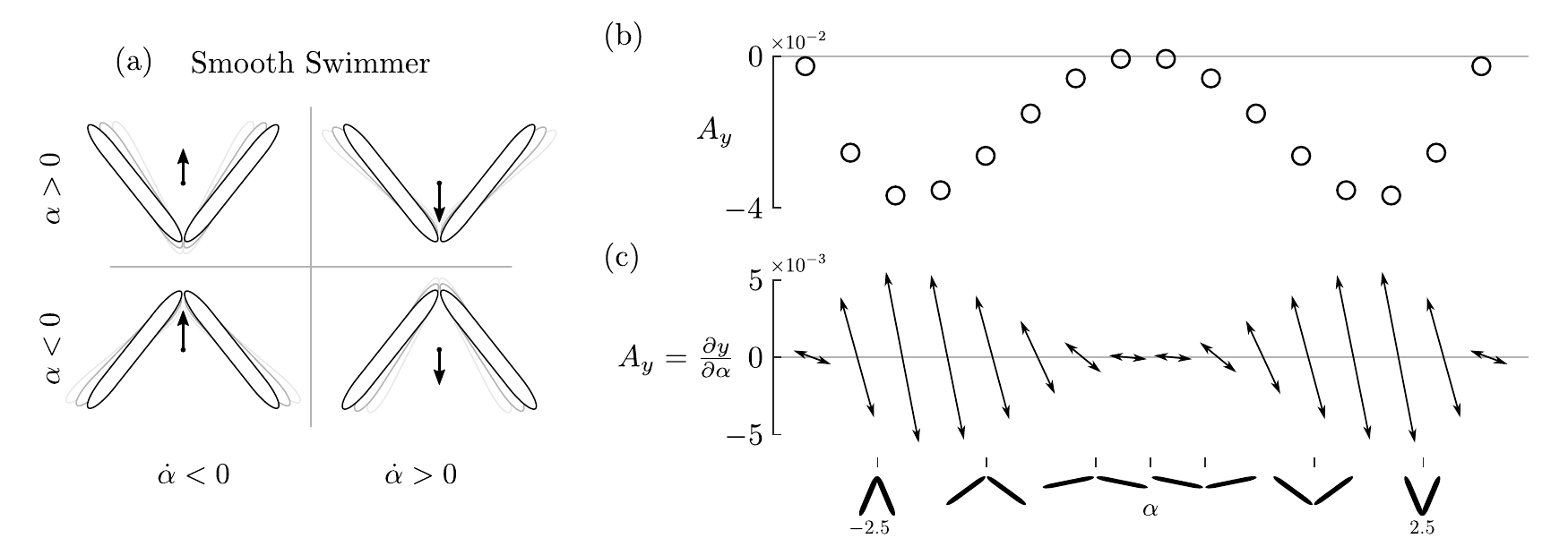}
    \caption{Two-link swimmer with no scales.
    (a) Qualitative motion in each quadrant of the shape-change, shape space $(\ShapeChange,\Shape)$. Motion-blur shows where the links were in the recent past. Arrows point from the center of mass in the direction of its motion, and show that only lateral motion is possible for this system.
    (b) The lateral component of the motility $\Motility_y$ as a function of $\Shape$. The motility is always negative, indicating that an increase in $\Shape$ will always lead to a decrease in $y$ and vice versa. The motility's magnitude peaks when the links are approximately perpendicular, meaning the most dramatic motion (for a given $\ShapeChange$) will occur around that shape.
    (c) $\Motility_y$ can be thought of as the slope $\frac{\partial y}{\partial \alpha}$. As $\Shape$ increases, $y$ follows the right arrow and because for this system that arrow always points down, $y$ decreases. Then, when $\Shape$ increases, it instead follows the left arrow and $y$ increases. Because the right and left arrows at each shape are of exactly opposite slope, all lateral motion cancels out when the swimmer returns to its original shape. No net displacement over a cycle is possible.}
    \label{fig:NoScalesSystem}
\end{figure*}

\section{Two-Link Swimmer Without Scales}
\label{sec:NoScales}

Previous research has emphasized swimmer systems with at least three links. Under reversible physics, systems with fewer links can not achieve net motion due to the Scallop Theorem\cite{Purcell:1977}, but with the scale-like lateral drag we examine in later sections, net motion can be achieved. This section will walk through the basics of geometric mechanics applied to the reversible, low Reynolds-number two-link (non-)swimmer case.
The plots and reasoning in this section mirror those in the next section so that the impact of having non-reversible lateral drag is as clear as possible.

\subsection{Finding The Body Velocity}
Geometric mechanics is the theory of using differential geometry to understand locomotion. It can help us explain why one gait is more effective than another, and what the potential locomotion performance is for a given system--- without having to simulate every possible gait.

We are modeling the environment as a highly viscous (i.e. low Reynolds number) fluid. Under this model, all transient effects die off so quickly that we can ignore them, and the swimmer's links are always moving at terminal velocity. Forces on the swimmer are always balanced. In this model, $F=0$ instead of the more familiar $F=ma$. This model can be used in systems where drag is much stronger than inertia, often used in microbiology.

We can use $F = 0$ as a Pfaffian constraint to obtain a linear differential equation for the body velocity $\BodyCirc$.
Modelling the drag on each link in its body frame as,
\begin{equation} \label{eq:DragMatrix}
    \DragMatrix_\LinkNumber =
    -\LinkLength
    \begin{bmatrix}
    \DragLon & 0 & 0\\
    0 & \DragLat & 0\\
    0 & 0 & \frac{1}{12} \DragLat \LinkLength^2 
    \end{bmatrix},
\end{equation}
we use a Jacobian $\Jacobian_\LinkNumber$, which takes the body configuration velocity and gets each link $i$'s velocity, then use the Jacobian again to pull back that link's drag into the body frame. Summing the drag force from both links, we get the net force $\BodyForce$,
\begin{equation}
    \BodyForce = \sum \Jacobian_\LinkNumber^T \DragMatrix_\LinkNumber \Jacobian_\LinkNumber,
\end{equation}
a $3 \times 4$ matrix that depends on the joint angle $\Shape$. Multiplying by the configuration velocity to get the vector net force on the system, constrained to zero by the Pfaffian constraint,
\begin{equation}
    \begin{bmatrix}
    0\\0\\0
    \end{bmatrix} =
    \BodyForce(\Shape)
    \begin{bmatrix}
    \BodyCirc \\ \ShapeChange
    \end{bmatrix}.
\end{equation}

We can separate $\BodyForce$ into two sub-matrices $\BodyForce_\BodyFrame$ ($3 \times 3$) and $\BodyForce_\Shape$ ($3 \times 1$),
\begin{equation}
    \begin{bmatrix}
    0\\0\\0
    \end{bmatrix}=
    \begin{bmatrix}
    {\BodyForce_\BodyFrame}^{3 \times 3} & {\BodyForce_\Shape}^{3 \times 1}
    \end{bmatrix}
    \begin{bmatrix}
    \BodyCirc \\ \ShapeChange
    \end{bmatrix},
\end{equation}
which can be understood respectively as the drag from translation and the drag from flapping if the swimmer were pinned in place.
Now, we multiply each sub-matrix by their corresponding components of the configuration velocity and subtract the $\BodyFrame$ term from both sides,
\begin{align}
    - \BodyForce_\BodyFrame \BodyCirc &=
    \BodyForce_\Shape \ShapeChange.
\intertext{
Finally, we multiply both sides by $-\BodyForce_\BodyFrame^{-1}$ on the left to obtain an equation for the body velocity,
}
    \BodyCirc &=
    -\BodyForce_\BodyFrame^{-1}
    \BodyForce_\Shape
    \ShapeChange.
\intertext{
The matrix multiplied by $\ShapeChange$ is called the motility Function $\Motility$,
}
    \BodyCirc &= \label{eq:Motility}
    \Motility \ShapeChange,
\end{align}
which depends only on $\Shape$ and contains all the information we need to understand the system's locomotion. For a more detailed derivation of $A$ including higher dimensions of $\Shape$, see \cite{Ramasamy:2019}.

\subsection{Understanding The Motility Function}
From \eqref{eq:Motility}, it follows that we can think of the motility as a slope for each component of the body configuration $\BodyFrame$ with respect to the shape configuration $\Shape$,
\begin{equation}
    \Motility_i(\Shape) = \frac{\partial \BodyFrame_i}{\partial \Shape},
\end{equation}
where $i$ is $x$, $y$, or $\theta$ in the body frame.

We can use the motility to explain why the swimmer will have no net displacement in a cycle.
The $x$ and $\theta$ components of the motility are not shown in Fig. \ref{fig:NoScalesSystem} because they are both zero for all $\Shape$, the swimmer doesn't move forward or turn, even within a gait.
Fig. \ref{fig:NoScalesSystem} shows $\Motility_y$ plotted as both its scalar values and as a field of slopes with $\partial \Shape$ held constant. Any slope $\frac{\partial \BodyFrame_y}{\partial \Shape}$ traversed in the positive $\Shape$ direction (shown by the right arrow on each slope) must also be traversed in the negative $\Shape$ direction (the left arrows). The change in $\BodyFrame_y$ from each direction then cancel exactly, resulting in motion within a gait cycle that exactly cancels when the swimmer returns to its original shape.
\begin{figure*}
    \centering
    \includegraphics[width=\textwidth]{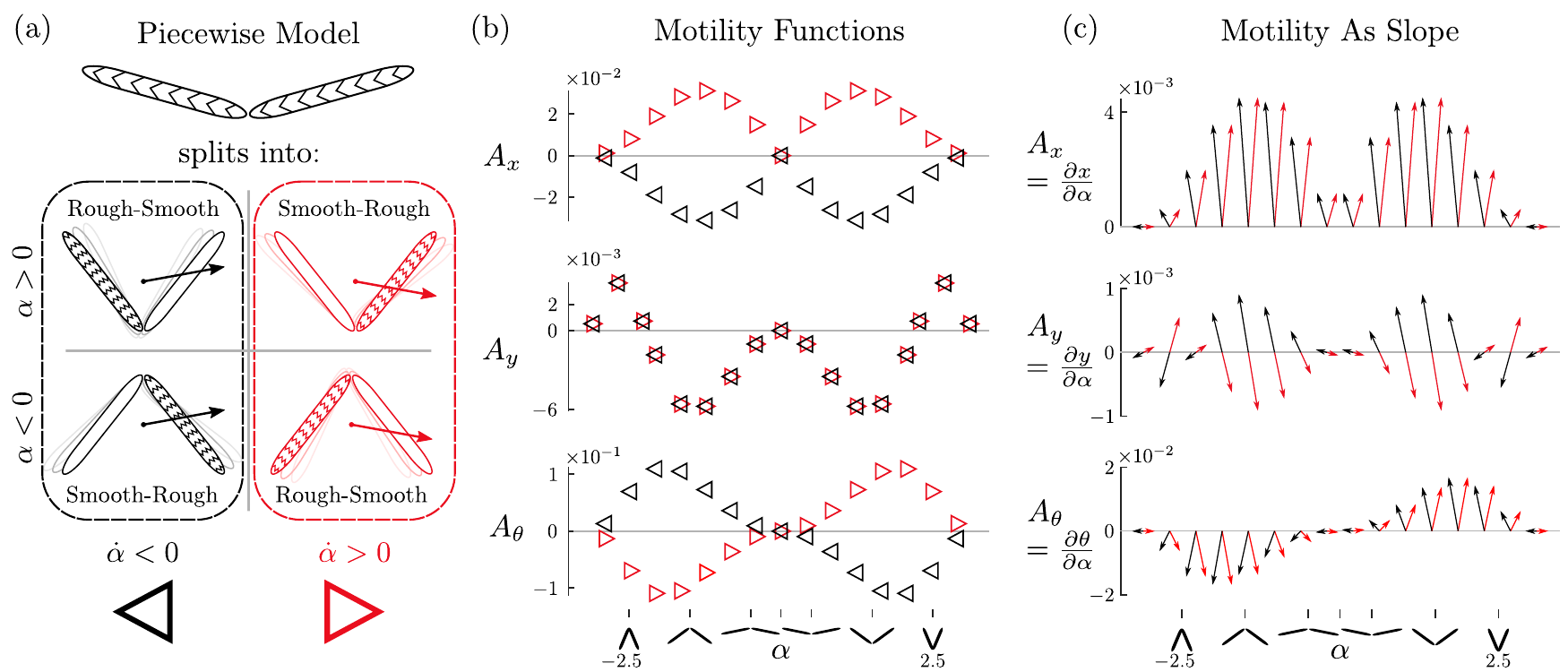}
    \caption{Two-link swimmer with scales.
    (a) Each scaled link is calculated as a rough or smooth link depending on its current longitudinal velocity, as described in \eqref{eq:uLon}. Only half of the four combinations (Rough-Smooth and Smooth-Rough, but not Smooth-Smooth or Rough-Rough) occur in the shape-change, shape space $(\ShapeChange, \Shape)$, and each quadrant in this space has only a single combination. In contrast to Fig. \ref{fig:NoScalesSystem}(a) which had no longitudinal motion, every quadrant in the scaled system has positive longitudinal motion. Quadrants are color-coded based on the piecewise motility functions defined in \eqref{eq:PlusMinusPiecewise}. Black and left-pointing triangles \MotilityLeft for $\MotilityMinus$ describe what happens when $\Shape$ is decreasing. Red and right-pointing triangles \MotilityRight for $\MotilityPlus$ describe what happens when $\Shape$ is increasing.
    (b) The motility functions for the scaled swimmer, split into $\MotilityMinus$ (\MotilityLeft) and $\MotilityPlus$ (\MotilityRight). The swimmer switches between the two motility functions when $\ShapeChange$ changes sign. The $x$ and $\theta$ components of $\MotilityMinus$ and $\MotilityPlus$ are equal in magnitude but opposite in sign, but the $y$ components are equal. The $x$ components are even while the $\theta$ components are odd, but from this kind of plot the significance of that distinction is obscured.
    (c) As in Fig. \ref{fig:NoScalesSystem}(c), the motility is plotted again as a field of slopes. For the scaled system, the left-pointing arrows have a slope according to $\MotilityMinus$ and the right-pointing arrows according to $\MotilityPlus$. $\Motility_y$ looks very similar to Fig. \ref{fig:NoScalesSystem}(c), with straight double-sided arrows. However $\Motility_x$ and $\Motility_\theta$ form V's due to the opposing signs of $\MotilityPlus$ and $\MotilityMinus$. When traversing a V from one direction, it will cause a motion that is then caused again in the same direction when traversing the V from the opposite direction. In fact, any amount of "bending" the slope due to $\MotilityPlus - \MotilityMinus \neq 0$ breaks reversibility and allows net displacement, as will be explained further in Section \ref{sec:ADiff}.
    }
    \label{fig:ScalesSystem}
\end{figure*}

\section{Two-Link Swimmer With Scales}
\label{sec:Scales}
Non-reversible drag is less straightforward to calculate, but common in nature. One example is the scales on snake bellies. Biologists believe that scales contribute to snake locomotion\cite{Hu:2009}, experiments on robots support this idea\cite{Branyan:2020}, but geometric mechanics does not yet have the tools to examine non-reversible drag.

To account for the non-reversible drag from scales as described in Figure \ref{fig:KindsOfDrag}, we modify the drag matrix $\DragMatrix_\LinkNumber$ from \eqref{eq:DragMatrix}. Now, breaking reversibility, the drag matrix depends on the longitudinal component of the link velocity $\LinkVelocity^x$,\footnote{When the longitudinal link velocity is zero, the drag is also zero regardless of which drag matrix is used, so it doesn't matter which is used for that case.}
\begin{equation}
    \DragLon = 
    \begin{cases}
    \FWRatio,& \text{when } \LinkVelocity^x \geq 0\\
    \BWRatio,& \text{when } \LinkVelocity^x < 0
    \end{cases}.
    \label{eq:uLon}
\end{equation}

However this new definition for $\DragLon$ breaks linearity. Using the derivation of the motility function from the previous section, we can model systems that have their experienced force only depend on the shape and shape change, but not on the body velocity. Because in this system the magnitude of friction depends on the link velocities, that won't work here. To fix the problem, we will split the scaled swimmer into separate systems that all have constant values of $\DragLon$ at each link.

\subsection{A Piecewise Model For Scales}
Instead of single system with non-reversible drag, we can consider reversible swimmers with each combination of smooth and rough links. For smooth links $\DragLon = \FWRatio$, and for rough links $\DragLon = \BWRatio$. There are four such two-link swimmers: Smooth-Smooth, Smooth-Rough, Rough-Smooth, and Rough-Rough. We are already familiar with Smooth-Smooth, as that was the system discussed in Section \ref{sec:NoScales}.

For a given $\BodyCirc$, the scaled swimmer will have drag matrices equal to those of one of these four reversible swimmers. Then, the scaled swimmer can be modeled as that reversible swimmer and use its motility function.

According to the directions of $\BodyCirc$ for the Smooth-Smooth swimmer in Fig. \ref{fig:NoScalesSystem}(a), there is no situation where both links have a forward link velocity. This makes sense because there is nothing to break the symmetry of the $\BodyFrame_x = 0$ axis. Because the Smooth-Smooth swimmer can't generate forward-forward link velocities, it (and, by a similar symmetry argument, the Rough-Rough swimmer) can not ever be consistent with the scaled swimmer.

Now we can restrict our attention to the Smooth-Rough and Rough-Smooth swimmers. Using a brute-force survey of the link velocities over the $(\ShapeChange, \Shape)$ space, we find that motion of the Smooth-Rough swimmer is consistent with that of the scaled swimmer for the $(\ShapeChange > 0,\ \Shape > 0)$ and $(\ShapeChange < 0,\ \Shape < 0)$ quadrants of the $(\ShapeChange, \Shape)$ space (when the swimmer is closing), and motion of the Rough-Smooth is consistent with that of the scaled swimmer for the other two quadrants (when the swimmer is opening), as illustrated in Fig. \ref{fig:ScalesSystem}(a).

We can thus model the scaled swimmer's motility function as a piecewise combination of the motility functions for Smooth-Rough $\MotilitySR$ and Rough-Smooth $\MotilityRS$,
\begin{equation}
    \Motility(\Shape,\ShapeChange) = 
    \begin{cases}
        \MotilitySR,& \frac{d|\Shape|}{dt} \geq 0
        \text{ (opening)}\\
        \MotilityRS,& \frac{d|\Shape|}{dt} < 0
        \text{ (closing)}
    \end{cases}.
    \label{eq:PiecewiseMotilityOne}
\end{equation}
(At the boundaries between quadrants $\MotilitySR=\MotilityRS$, so the decision of where to put the $=$ is arbitrary.)

In the next section, it will be helpful to define a new pair of motility functions $\MotilityPlus$ and $\MotilityMinus$ as piecewise combinations of $\MotilitySR$ and $\MotilityRS$ based on the sign of $\ShapeChange$:
\begin{align}
    \Motility(\Shape, \ShapeChange) &=
    \begin{cases}
        \MotilityPlus,& \ShapeChange \geq 0\\
        \MotilityMinus,& \ShapeChange \leq 0
    \end{cases},
    \label{eq:PlusMinusPiecewise}
    \\
%
\text{where }
    \MotilityPlus(\Shape) &=
    \begin{cases}
        \MotilitySR,& \Shape \geq 0\\
        \MotilityRS,& \Shape \leq 0
    \end{cases}
    \\
\text{and }
    \MotilityMinus(\Shape) &=
    \begin{cases}
        \MotilityRS,& \Shape \geq 0\\
        \MotilitySR,& \Shape \leq 0
    \end{cases}.
\end{align}
Fig. \ref{fig:ScalesSystem} splits the scaled swimmer's motility function according to this definition, using red for $\MotilityPlus$ and black for $\MotilityMinus$. The meaning of these piecewise motility functions will be explained further in the next section.

Now we have a motility function that depends on $\Shape$ and $\ShapeChange$, and not explicitly on $\BodyCirc$. Calculating the net displacement of a single gait is again possible by numerically solving \eqref{eq:Motility}.

\subsection{Understanding The Piecewise Motility Function}
Figure \ref{fig:ScalesSystem}b shows the $\Motility$ broken down into its $x$, $y$, and $\theta$ components. On each of these subplots, the red triangles \MotilityRight pointing in the direction of increasing $\Shape$ is $\MotilityPlus$, and the black triangles \MotilityLeft pointing towards decreasing $\Shape$ is $\MotilityMinus$.

As the swimmer increases its joint angle $\Shape$, its resulting body velocity scales according to the value of $\MotilityPlus$ at its current $\Shape$ value. As it decreases its joint angle $\Shape$, the same thing happens but according to the current value of $\MotilityMinus$.
A more mathematically appropriate way to view the motility function is as a field of slopes, one at each combination of shape variables (in a single-jointed snake, that's one at each value of $\Shape$). The value of $\Motility_x$ (or $\Motility_y$, $\Motility_\theta$) is not a height on some axis, but the slope of the function relating $\Shape$ to $\BodyCirc$ at that $\Shape$ value. For one shape variable, we can get away with viewing it as integration, but that perspective doesn't generalize correctly to a higher-dimensional shape-space.
The $\Motility$ value at the zero axis determines the slope of each line segment. The width of the line segments, the change in $\Shape$ over the slope, is constant for each line segment. This way, the magnitude of the slope is clearly visible from the height reached by the line segment.

Because $\MotilityPlus$ is only every used for increasing $\Shape$ and $\MotilityMinus$ for decreasing $\Shape$, the part of the red line in the negative direction and the black line in the positive direction are never used.
Instead of two separate slopes, we can consider each pair as a single bent ramp. A system without scales would have one straight line for all of its ramps. This system has scales, which bend its one straight line.
For example on the $\Motility_x$ subplot, you can see at a glance that a gait that always remains in the negative $\Shape$ region will have positive $\BodyCirc_x$:
Starting at any non-zero $\Shape$ and increasing $\Shape$ will push you up the red $\MotilityPlus$ ramp: $\MotilityPlus_x$ is positive for all non-zero $\Shape$, so the slope is positive. This means that $\BodyCirc_x$ will be positive during this motion. Decreasing $\Shape$ puts us on the black $\MotilityMinus$ ramp. The value of $\MotilityMinus_x$ is negative for all non-zero $\Shape$, but we're going backwards on the ramp, $\BodyCirc_x$ is still positive.

\begin{figure*}
    \centering
    \includegraphics[width=\textwidth]{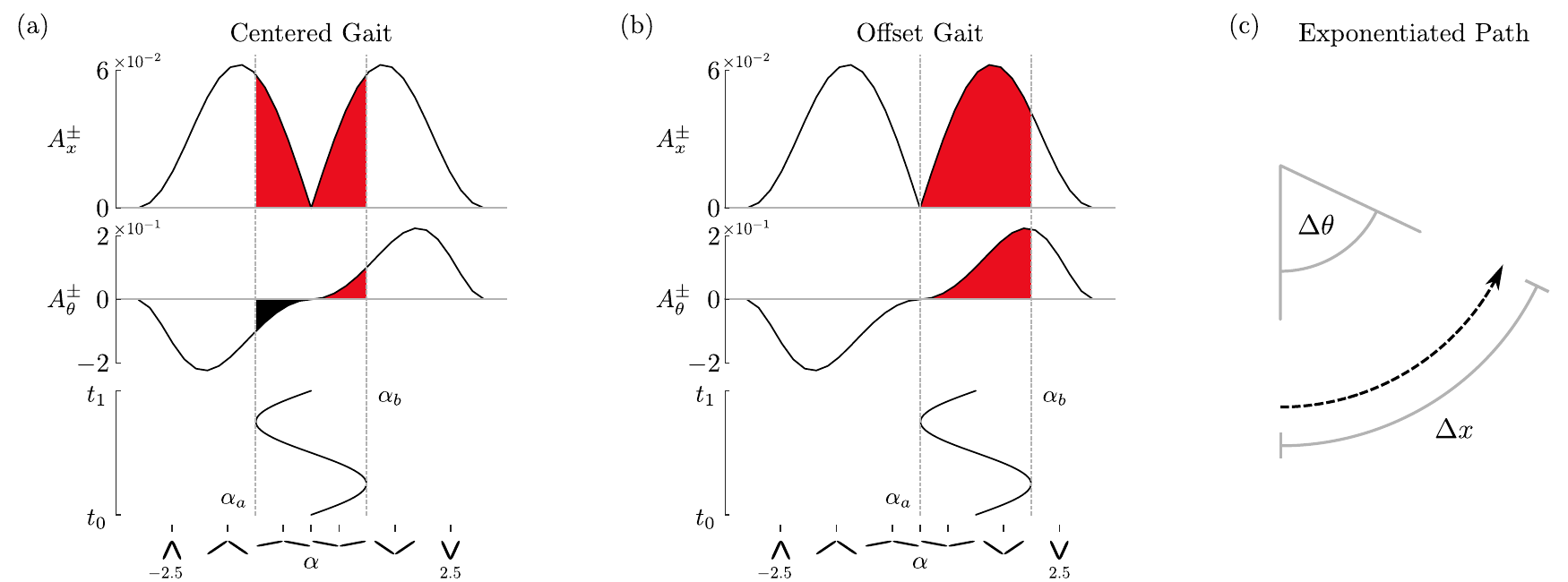}
    \caption{Integrating over the difference in motility functions $\MotilityDiff$ gives an approximation of gait displacement.
    (a) A sinusoidal gait centered around $\Shape = 0$ ranging from $\Shape_a$ to $\Shape_b = -\Shape_a$. The areas under $\MotilityDiff_x$ and $\MotilityDiff_\theta$ between $\Shape_a$ and $\Shape_b$ predict the displacement this gait will produce. Because $\MotilityDiff_\theta$ is odd, we expect this gait to have no net rotation. The area under $\MotilityDiff_x$ is positive, so we expect the swimmer to move forward in a straight line.
    (b) A sinusoidal gait ranging from $\Shape_a = 0$ to $\Shape_b > 0$. From $\Shape_a$ to $\Shape_b$, the areas under $\MotilityDiff_x$ and $\MotilityDiff_\theta$ are both positive. We expect this gait to have both a net rotation and net longitudinal motion.
    (c) The change in $x$ and $\theta$ predicted by integrating $\MotilityDiff$ are in the body frame $\BodyFrame$. However $\BodyFrame$ rotates as $\theta$ changes. We would like the displacement in the world frame. Because we have abstracted away time dependence, we assume the displacement is evenly distributed, tracing out the arc of the dashed line.
    }
    \label{fig:APlusMinus}
\end{figure*}

\section{Approximating Displacement With Integration}
\label{sec:ADiff}
In Section \ref{sec:Scales}, we were able to see how $\Shape$ and $\ShapeChange$ result in a $\BodyCirc$. By following one cycle of a gait $\Gait$ from times $\Time_0$ to $\Time_1$, we could computationally solve \eqref{eq:Motility} to get the net displacement $\Delta \BodyFrame$ for a single gait, updating $\BodyFrame$ at each point in time.
One of the goals of geometric mechanics is to uncover patterns about the system as a whole, not just a specific gait. In this section we will develop
a new way to understand the piecewise motility function. Then,
predicting net displacement will be as simple as measuring the area under a curve.

The net displacement $\Delta \BodyFrame$ depends on the intermediate transformations on the body frame during a specific gait. To instead examine the system as a whole, we must make the approximation that the displacement is instead accumulated in a single frame $\BodyFrame'$, which must then be exponentiated to account for the changing frame,
\begin{equation}
    \Delta \BodyFrame \simeq \exp{\Delta \BodyFrame'}.
\end{equation}
An illustration of this equation is shown in Fig. \ref{fig:APlusMinus}(c). Now, the pre-exponentiated displacement can be stated with an integral of $\BodyCirc$ over time,
\begin{align}
    \Delta \BodyPrime &=
        \int_{\Time_0}^{\Time_1}
        \BodyCirc \, d\Time.
        \\
\intertext{Using \eqref{eq:Motility} to get $\BodyFrame$ in terms of $\ShapeChange$,}
    \Delta \BodyPrime &=
        \int_{\Time_0}^{\Time_1}
        \Motility \ShapeChange \, d\Time,
        \\
\intertext{which according to the definition of $\ShapeChange$ is}
    \Delta \BodyPrime &=
        \int_{\Time_0}^{\Time_1}
        \Motility \frac{d\Shape}{d\Time} \, d\Time.
        \\
\intertext{The $d\Time$ terms cancel, leaving us with an integral of $\Motility$ over the $\Shape$ values of one cycle of $\Gait$,}
    \Delta \BodyPrime &=
        \oint_{\Gait}
        \Motility \, d\Shape.
    \label{eq:MotilityIntegral}
\end{align}
Because a cycle of a gait must be a closed loop and we only have one degree of freedom, the swimmer must cross values of $\Shape$ an even number of times. For any time it crosses from below ($\ShapeChange > 0$), it must cross again from above ($\ShapeChange < 0$) in order to return to the starting shape.\footnote{Minima and maxima of the gait may be reached an odd number of times, but are not crossed.} We can then split the integral over $\Gait$ into integrals over the parts of $\Gait$ with different signs of $\ShapeChange$. Using the piecewise motility function defined in \eqref{eq:PlusMinusPiecewise},
\begin{align}
    \Delta \BodyPrime &=
        \int_{\Gait^+}
        \MotilityPlus \, d\Shape
        +
        \int_{\Gait^-}
        \MotilityMinus \, d\Shape.
        \\
\intertext{Because each shape in $\Gait^+$ has a counterpart in $\Gait^-$ going the opposite direction, these integrals can be rewritten as one,}
    \Delta \BodyPrime &=
        \int_{\Gait}
        \left(\MotilityPlus - \MotilityMinus\right) \, d\Shape,
\end{align}
where we define the open integral (as opposed to the closed integral in \eqref{eq:MotilityIntegral}) over a gait $\Gait$ as its integral over $\Gait^+$.
We define the difference of motilities $\MotilityDiff$ to be,
\begin{equation}
    \MotilityDiff = \MotilityPlus - \MotilityMinus,
\end{equation}
and now the integral is simply
\begin{align}
    \Delta \BodyPrime &=
        \int_{\Gait}
        \MotilityDiff \, d\Shape.
        \\
\intertext{ For gaits like those in Fig. \ref{fig:APlusMinus} where no value of $\Shape$ is covered more than twice, the bounds of the integral become a simple range,}
    \Delta \BodyPrime &=
        \int_{\Shape_\text{min}}^{\Shape_\text{max}}
        \MotilityDiff \, d\Shape,
\end{align}
where $\Shape_\text{min}$ and $\Shape_\text{max}$ are the minimum and maximum values reached over the gait.

Fig. \ref{fig:APlusMinus} shows $\MotilityDiff_x$ and $\MotilityDiff_\theta$. $\MotilityDiff_y$ is not shown because it is zero for all $\Shape$.\footnote{The only source of $y$ displacement is from combinations of $x$ and $\theta$ motion over the course of a gait.}
Two example sine-wave gaits are shown. One centered around a straight shape, and one centered around a bent shape. It is simple to predict the displacement produced by these gaits. Look at the area under the curves to find $\BodyPrime$, then remember to exponentiate as shown in Fig. \ref{fig:APlusMinus}(c).
\section{Conclusion}

As a starting point, we used established geometric mechanics techniques to model the reversible, no-scales swimmer. We modeled the non-reversible, scaled swimmer as a combination of swimmers without scales, but with links of differing roughness. We observed that switching between the behaviors of these reversible systems allowed the scaled swimmer to behave in a non-reversible manner, moving forward regardless of shape or shape change. Finally, we conveyed this motion succinctly in Fig. \ref{fig:APlusMinus} with $\MotilityDiff$, the difference between two reversible motility functions.

We now have the ability to understand how scales impact the locomotion of the two-link swimmer without needing to limit our view to specific gaits. Our piecewise model allowed us to take the difference between opening and closing motions of the joint, pairing opposing portions of any gait.

The resulting quantity over the shape space is reminiscent of the curl and lie bracket operations used effectively for reversible systems with more than one joint.

In future work, we will combine the techniques established in this paper with the curl and lie bracket techniques of reversible systems to extend our understanding of scales to a higher-dimensional shape space.

Non-reversible systems such as those with scales impose a Finsler (as opposed to Reimannian) metric on the configuration space, where the distance from point A to B is not the same as the distance from point B to A. Thus by investigating the underlying geometric properties of a non-reversible form of locomotion, we are joining \cite{Ratliff:2021,Huang:2020,Xie:2021} in creating a new avenue for applied mathematics in robotics.
\section*{Acknowledgment}
This work was supported in part by the National Science Foundation under grants 1653220 and 1826446

\bibliographystyle{plain}
\bibliography{bibliography/bibliography.bib, bibliography/rossbib.bib}

\vspace{12pt}

\end{document}